\documentclass[sigconf, authorversion, nonacm]{acmart}

\AtBeginDocument{%
  }

\setcopyright{none}

\usepackage{geometry}
\usepackage{graphicx} 

\usepackage{bm}
\usepackage{amsmath}
\usepackage{xcolor}
\usepackage{float}
\usepackage{xspace}
\usepackage{mathtools}
\usepackage{natbib} 
\usepackage{enumitem}

\usepackage{algorithmicx}
\usepackage{algorithm}
\usepackage{algpseudocode}

\usepackage{multirow} 
\usepackage{booktabs}
\usepackage{arydshln}

\usepackage{tikz}
\usepackage{pgfplots}
\pgfplotsset{compat=1.3}
\usepackage{subcaption}
\usepackage{colortbl}
\usepackage{makecell}

\usepackage{adjustbox}

\newcommand{\defn}{\coloneqq}

\newcommand{\cA}{\mathcal{A}}

\newcommand{\cD}{\mathcal{D}}
\newcommand{\cE}{\mathcal{E}}

\newcommand{\cL}{\mathcal{L}}

\newcommand{\cS}{{\mathcal{S}}}

\usepackage{scalerel,stackengine}
\stackMath
\newcommand\reallywidehat[1]{%
\savestack{\tmpbox}{\stretchto{%
  \scaleto{%
    \scalerel*[\widthof{\ensuremath{#1}}]{\kern-.6pt\bigwedge\kern-.6pt}%
    {\rule[-\textheight/2]{1ex}{\textheight}}
  }{\textheight}%
}{0.5ex}}%
\stackon[1pt]{#1}{\tmpbox}%
}
\newcommand\reallywidecheck[1]{%
\savestack{\tmpbox}{\stretchto{%
  \scaleto{
    \scalerel*[\widthof{\ensuremath{#1}}]{\kern-.6pt\bigwedge\kern-.6pt}%
    {\rule[-\textheight/2]{1ex}{\textheight}}
  }{\textheight}%
}{0.5ex}}%
\stackon[1pt]{#1}{\scalebox{-1}{\tmpbox}}%
}

\definecolor{jiw}{RGB}{10,148,15}

\DeclareMathOperator*{\argmax}{arg\,max}
\DeclareMathOperator*{\argmin}{arg\,min}

\newcommand{\algoname}{{\sf LAAC}\xspace}

\begin{document}

\title[LLM-Enhanced RL for Diverse and Novel Recommendations]{Large Language Model-Enhanced Reinforcement Learning\\ for Diverse and Novel Recommendations}

\author{Jiin Woo}
\email{jiinw@andrew.cmu.edu}
\affiliation{%
  \institution{Carnegie Mellon University}
    \country{}
}

\author{Alireza Bagheri Garakani}
\email{alirezg@amazon.com}
\affiliation{%
  \institution{Amazon}
  \country{}
}

\author{Tianchen Zhou}
\email{tiancz@amazon.com}
\affiliation{%
  \institution{Amazon}
  \country{}
}

\author{Zhishen Huang}
\email{hzs@amazon.com}
\affiliation{%
  \institution{Amazon}
  \country{}
}

\author{Yan Gao}
\email{yanngao@amazon.com}
\affiliation{%
  \institution{Amazon}
  \country{}
}










\begin{abstract}
In recommendation systems, diversity and novelty are essential for capturing varied user preferences and encouraging exploration, yet many systems prioritize click relevance. While reinforcement learning (RL) has been explored to improve diversity, it often depends on random exploration that may not align with user interests. We propose LAAC (LLM-guided Adversarial Actor Critic), a novel method that leverages large language models (LLMs) as reference policies to suggest novel items, while training a lightweight policy to refine these suggestions using system-specific data. The method formulates training as a bilevel optimization between actor and critic networks, enabling the critic to selectively favor promising novel actions and the actor to improve its policy beyond LLM recommendations.
To mitigate overestimation of unreliable LLM suggestions, we apply regularization that anchors critic values for unexplored items close to well-estimated dataset actions. Experiments on real-world datasets show that LAAC outperforms existing baselines in diversity, novelty, and accuracy, while remaining robust on imbalanced data, effectively integrating LLM knowledge without expensive fine-tuning.
\end{abstract}

\maketitle

\section{Introduction}
In recommendation systems (RS), diversity and novelty are crucial factors to address diverse user preferences and promote exploration of new interests.
Many search and recommendation systems tend to prioritize click relevance, which often limits diversity, especially when data or user experiences are limited. Recognizing that reinforcement learning (RL) can offer personalized and dynamic recommendations by prioritizing long-term user satisfaction, several studies have investigated RL methods, enhancing diversity in recommendations \cite{stamenkovic2022choosing,liu2022diversity,zheng2018drn} by encouraging random exploration or introducing explicit rewards for diversity. Such approaches may not consistently yield optimal results, as randomly selected items may not cater to users' needs.

A more effective approach is to leverage prior knowledge to explore novel items selectively. Large language models (LLMs), trained on enormous corpus of data, can be a reliable source of novel items beyond the system’s limited user experiences. To utilize LLMs to enhance RS, some works have used LLMs as recommenders by generating recommendations directly from LLMs with prompts \cite{he2023large} or fine-tuning LLMs \cite{bao2023tallrec}, but these approaches incur significant computational costs for inference. Meanwhile, recent research has suggested to use LLMs as an environment component to augment limited training data \cite{wang2024reinforcement}, reducing inference-time computational costs. This approach still requires significant computational resources for LLM training and risks losing the diverse user preferences learned during pretraining when models are retrained on specific datasets. Therefore, it is crucial to strategically integrate LLM suggestions into RL training without costly fine-tuning, to ensure fast adaptation to new data, scalability, and preservation of diverse user preferences learned during pretraining.

To address this issue, we investigate an RL method that uses LLM as a reference policy rather than retraining it. We can leverage its ability to suggest potentially appealing items and train a separate, lightweight policy to refine and align these recommendations using datasets collected from target systems. 
However, using LLMs’ suggestions without additional training can be risky, as not all recommendations of LLMs meet user expectations in the target systems, potentially compromising user satisfaction. Therefore, it's crucial to be selectively optimistic on LLMs' suggestions likely to yield high rewards, while also retaining popular items proven effective in datasets. To this end, we formulate the RL problem as a two-player game between a policy and a critic function \cite{cheng2022adversarially,bhardwaj2023adversarial}, where the policy seeks to learn actions that outperform the LLMs' suggestions, while the critic aims to learn realistic value functions that favor those suggestions.
The adversarial dynamics between the policy and critic prevent both greedy exploitation of popular items in the datasets and blind optimism towards LLM recommendations, enabling the selective integration of novel items outside the dataset's coverage.

In this paper, we present an LLM-guided adversarial actor-critic training method (\algoname) that iteratively refines the actor and critic networks based on LLM recommendations and datasets, optimizing the adversarial objectives against one another. 
To prevent overestimation on unreliable LLM suggestions, we incorporate novel regularizations that ensure the critic values for unexplored items remain close to those for actions reliably estimated from the dataset.
As a result, the algorithm develops a balanced policy that recommends both popular items in the dataset and novel items suggested by LLMs.
We summarize our main contributions as follows:
\begin{itemize}[leftmargin=*]
\item We propose an efficient actor-critic training method that effectively integrates LLMs to promote diversity and novelty in RS through adversarial RL training, without costly LLM training.
\item We conduct experiments using real-world datasets (MovieLens), demonstrating that our algorithm improves diversity, novelty, and accuracy over baseline models, while also proving robust on imbalanced data.
\item We analyze trade-offs between accuracy and novelty when choosing regularization parameters ($\alpha$ and $\beta$) by testing various values.
\end{itemize}

\subsection{Related work}

\subsubsection{Reinforcement learning for diversity and novelty in recommendation systems}
In RS, diversity and novelty are key factors for enhancing user experience and uncovering hidden preferences. Recent studies have explored RL algorithms to deliver more diverse and dynamic recommendations, focusing on long-term user satisfaction. To enhance diversity, \cite{zheng2018drn} examined exploration-exploitation strategies in RL by randomly selecting item candidates from the vicinity of the currently recommended item. To maximize diversity in recommendation results, while preserving relevance, \cite{liu2022diversity} integrated a determinantal point process model with the deep deterministic policy gradient algorithm. Lastly, \cite{stamenkovic2022choosing} introduced a scalarized multi-objective RL algorithm that optimizes three key reward objectives: accuracy, diversity, and novelty. 

\subsubsection{Large language models for recommendation systems}
LLMs~\cite{zhao2023survey,liu2023pre,min2023recent}, pre-trained on large natural language datasets, demonstrate improved transfer capabilities and are receiving growing interest in the field of recommendation systems~\cite{chen2023large,lin2023can,fan2023recommender}.
To harness LLMs for recommendation tasks, some studies have suggested using LLMs as recommenders directly, generating suggestions with proper prompts in a zero-shot manner~\cite{he2023large}. Additionally, to align LLMs with specific recommendation systems, other research has proposed retraining these models to serve as new recommenders via fine-tuning~\cite{min2023recent} or prompt-based tuning~\cite{liu2023pre,ouyang2022training}. Recent studies~\citep{hou2022towards,wang2022transrec} have explored training recommenders by analyzing user-item interactions alongside item features extracted from BERT~\cite{devlin2018bert}, yielding promising results for cross-domain recommendations. Furthermore, \cite{bao2023tallrec} efficiently fine-tunes LLaMA-7B~\cite{touvron2023llama} using LoRA adapters~\cite{hu2021lora} and an instruction prompt that incorporates item text descriptions to facilitate few-shot recommendations. However, these methods for employing LLMs as recommenders often entail significant time and computational costs during both training and inference. Recently, \cite{wang2024reinforcement} has proposed a method that uses LLMs as an environment component within an RL framework, with the objective of augmenting the performance of existing recommenders, which reduces computational costs at inference time. Nevertheless, this approach still requires considerable computation for LLM training to function as an environmental simulator, and it risks losing the diverse user preferences learned during pre-training when LLMs are retrained on specific datasets.

\section{Method}

\subsection{Preliminary}
\subsubsection{RL formulation} In RL, sequential RS can be framed as a Markov Decision Process (MDP), where an agent interacts with users by sequentially recommending items to maximize clicks or ratings.
The MDP is characterized by \( M = (\mathcal{S}, \mathcal{A}, P, R, \rho_0, \gamma) \), where \( \mathcal{S} \) represents the user's state, \( \mathcal{A} \) is a set of candidate action items, 
and the transition probability \( P: \mathcal{S} \times \mathcal{A} \times \mathcal{S} \rightarrow [0,1] \) defines \( p(s'|s, a) \), indicating the likelihood of transitioning from state \( s \) to state \( s' \) when action \( a \) is taken. The reward function \( r: \mathcal{S} \times \mathcal{A} \mapsto \mathbb{R} \) specifies that \( r(s, a) \) gives a rating or click from users for item \( a \) in state \( s \). \( \rho_0 \) denotes the initial state distribution and \( \gamma \in [0,1] \) is the discount factor.
A policy $\pi$ denotes an action-selection rule where $\pi(a| s)$ represents the probability of taking action $a$ in state $s$. 
For a given policy $\pi,$ the {\em value function} $V^{\pi}: \cS \rightarrow \mathbb{R}$, which measures the expected discounted cumulative reward from an initial state $s$, defined as
\begin{align*}  
   V^{\pi}(s) \defn \mathbb{E} \left[ \sum_{t=0}^{\infty} \gamma^t r(s_t,a_t ) \,\big|\, s_0 =s \right] . 
\end{align*}
Similarly, the state-action value function (i.e., {\em Q-function}) $Q^{\pi}: \cS \times \cA \rightarrow \mathbb{R}$, which measures the expected discounted cumulative reward from an initial state-action pair $(s,a)$, defined as   
\begin{align*}
  Q^{\pi}(s,a) \defn r(s ,a  )  + \mathbb{E} \left[  \sum_{t=1}^{\infty} \gamma^t r(s_t,a_t ) \,\big|\, s_0 =s, a_0 = a \right].
\end{align*}
Here, the expectation is taken with respect to the randomness of the trajectory $\{s_t,a_t,r_t\}_{t=0}^{\infty}$, sampled based on the transition probability (i.e., $s_{t+1}\sim P(\cdot | s_t, a_t)$) and the policy $\pi$ (i.e., $a_t \sim \pi(\cdot|s_t)$) for any $t \ge 0$. 
In this setting, the goal of the agent is to learn an optimal policy $\pi^{\star}$ that maximizes expected total rewards (ratings), i.e., $\pi^{\star} =\argmax_{\pi} \mathbb{E}_{s \sim \rho_0} \left[V^{\pi}(s) \right]$.

\subsubsection{Limited datasets and a reference policy} When the underlying MDP is unknown, the optimal policy can be trained from a dataset $\cD$ composed of transition samples $(s, a, r, s')$ collected from past user interactions, which provide the user response $r=r(s,a)$ and the user state transition $s' \sim P(\cdot|s,a)$ for an interacted item $a$ at state $s$.
However, the vast item space in RS makes exhaustive exploration impractical, leading the agent -- trained on limited datasets -- to focus on a narrow selection of popular items in datasets, which reduces diversity and overlooks potentially appealing yet unexplored items.
One effective approach to address the limitation of datasets with insufficient coverage of good items is to utilize a {\em reference policy}. In many RL studies, leveraging baseline/reference policies or experts to regularize RL training has been shown to enhance sample efficiency and training stability \cite{schulman2017proximal,wu2019behavior,bhardwaj2023adversarial,jaques2020human,stiennon2020learning}. Especially, when datasets lack sufficient observations to determine the optimal actions, these reference policies can offer safe alternatives. With the support of a reliable reference policy, agents can effectively learn novel actions beyond the dataset with minimal risk.

\subsubsection{Robust policy improvement via adversarial optimization}
A game-theoretic formulation for RL has been introduced in \cite{cheng2022adversarially} to safely improve a behavior policy (the policy used to collect training data), formulated as a bilevel optimization problem:
$$\max_\pi g(\pi,f^{\pi}), ~\text{s.t.}~f^{\pi} \in \argmin_f h(\pi,f),$$ where the two competing players are a policy $\pi$ (action selector) and a critic $f$ (value estimator), with $g$ and $h$ as their respective objective functions.
They proposed an actor-critic algorithm (ATAC) that solves this optimization through adversarial training, where the policy and critic compete against each other. The policy $\pi$ improves based on values predicted by the critic $f^\pi$, which deliberately provides pessimistic evaluations of $\pi$ compared to the original behavior policy used for dataset collection. This adversarial training setup guarantees that the resulting policy provably outperforms the behavior policy across a wide range of hyperparameter choices, though it restricts the policy to only recommend actions that were observed in the training dataset. Recently, \cite{bhardwaj2023adversarial} extended this adversarial optimization framework to incorporate an arbitrary reference policy (which can be different from the behavior policy) in model-based RL settings.
This extension enables developing an improved policy based on the reference policy, allowing actions beyond the available data support.

\begin{figure}[t]
    \centering
    \includegraphics[width=\columnwidth]{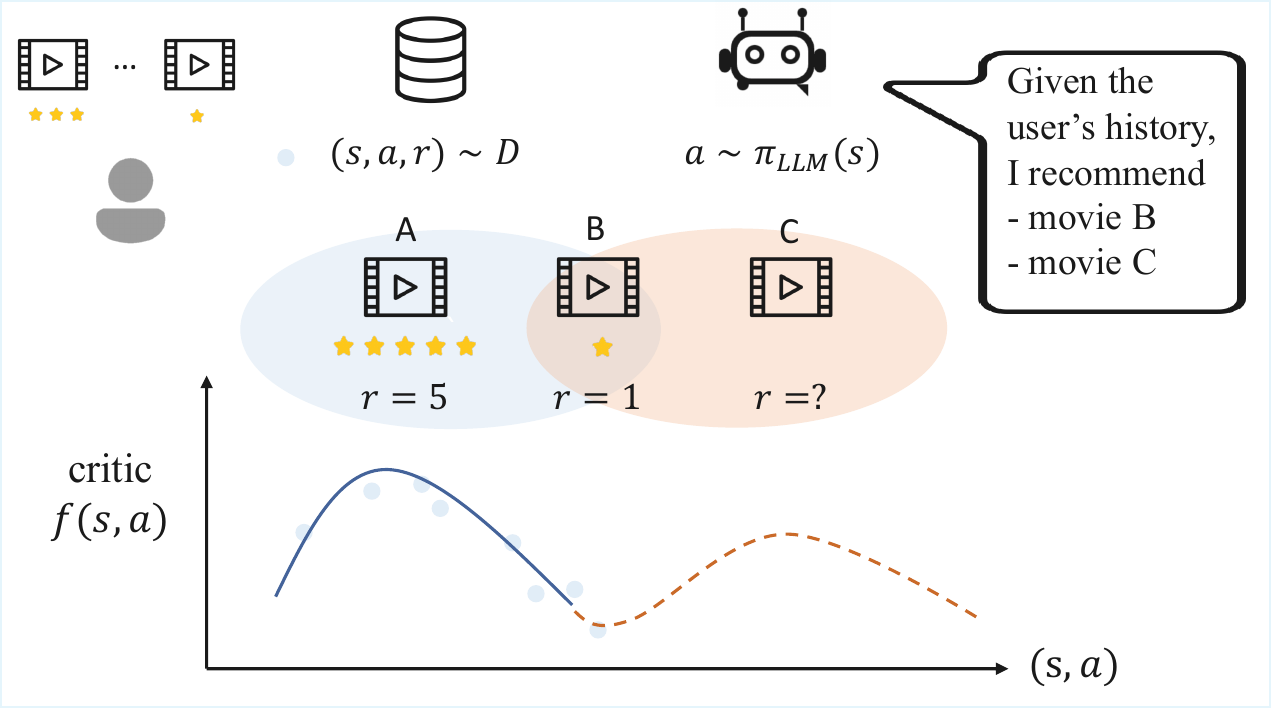}
    \caption{Illustration of the critic function $f$ in movie recommendation scenarios. The critic estimates realistic values for movies observed in dataset $\mathcal{D}$ based on actual rewards, while maintaining optimistic estimates for novel movies recommended by $\pi_{\textrm{LLM}}$ but absent from the dataset.}
    \label{fig:reward_hypothesis}
\end{figure}

\subsection{Adversarial optimization guided by LLMs}

In this paper, we explore an RL approach that enhances the diversity of a policy through adversarial optimization \cite{cheng2022adversarially,bhardwaj2023adversarial}, using LLMs as a reference policy to suggest novel and potentially appealing items beyond the dataset, thereby improving diversity while avoiding excessive exploration. Given their proven effectiveness in recommendation tasks \cite{deldjoo2024review,he2023large}, LLMs serve as a strong candidate for a reliable reference policy in RS. Here, we denote a reference policy generated from LLMs as $\pi_{\textrm{LLM}}$, where $\pi_{\textrm{LLM}}(a|s)$ indicates the probability of the LLM suggesting action $a$ for state $s$. Our goal is to obtain a balanced policy that recommends a diverse mix of popular and novel items by carefully managing the trade-off between exploring new items from the LLM policy $\pi_{\textrm{LLM}}$ and exploiting high-reward popular items from the dataset $\cD$.

To integrate novel items from the LLM policy while retaining popular items from the dataset, a policy should be trained to select the novel items only when data is insufficient and optimal actions are uncertain, while prioritizing high-reward items when enough observations are available.
To this end, inspired by the game-theoretic formulation of RL \cite{cheng2022adversarially,bhardwaj2023adversarial}, we formulate the problem as the following minimiax optimization:
\begin{align} \label{eq:llm_adv_opt}
    &\hat{\pi} = \argmax_{\pi} E_{(s,a,r,s')\sim\cD} \Big [ f(s, \pi) - f (s, \pi_{\textrm{LLM}}) \Big ] \cr
  & \text{s.t. }  f = \argmin_{f}  E_{(s,a,r,s')\sim\cD} \Big [ f(s, \pi) - f (s, \pi_{\textrm{LLM}}) \Big ]  \cr
    &\qquad\qquad\quad\quad\quad  + \alpha \cE_{\textrm{g}}(f,\pi_{\textrm{LLM}}) + \beta \cE_{\textrm{td}}(f,\pi)
\end{align}
where
$\alpha, ~\beta \ge 0$ are hyper parameters for the following losses:
\begin{align*}
    \cE_{\textrm{g}}(f,\pi_{\textrm{LLM}}) &\defn E_{(s,a,r,s') \sim \cD } \Big [ ( f(s,a) - f(s, \pi_{\textrm{LLM}} ))^2  \Big ] \\
    \cE_{\textrm{td}}(f,\pi) &\defn E_{(s,a,r,s') \sim \cD} \Big [ ( f(s,a) - r - \gamma f(s', \pi) )^2 \Big ].
\end{align*}
Here, $f(s,\pi) = E_{a \sim \pi(\cdot|s)}[f(s,a)]$, where the critic function $f(s,a)$ represents the estimated expected reward for item $a$ at state $s$, approximating the Q-function $Q(s,a)$.
This optimization can be viewed as a two-player game, where one player refines a policy $\pi$ in relation to the LLM policy $\pi_{\textrm{LLM}}$ based on a given critic value $f$, while the other player adversarially updates the critic $f$ to encourage optimism towards novel actions suggested by $\pi_{\textrm{LLM}}$ over $\pi$ as illustrated in Figure~\ref{fig:reward_hypothesis}. Ultimately, this process optimizes the policy $\pi$ for the worst-case performance ($f$) inferred from both the dataset $\cD$ and the LLM policy $\pi_{\textrm{LLM}}$.
Notably, the optimization process only necessitates updating the policy $\pi$ and the critic $f$, without the need for training of LLMs, which can be computationally intensive and time-consuming. In this manner, we can guide the policy to learn only effective actions from $\pi_{\textrm{LLM}}$ consistent with the provided dataset $\cD$, even if $\pi_{\textrm{LLM}}$ it is not perfectly aligned with the target system. 

\begin{algorithm}[t]
  \caption{\algoname (LLM-guided Adversarial Actor Critic)}
\label{alg:latac}
{\bfseries Input:} Dataset $\cD$, discount factor $\gamma \in [0,1]$, constants $\alpha, ~\beta\geq 0$, learning rates $\eta_{\textrm{critic}},~\eta_{\textrm{actor}}$, candidate size $n_c$, response size $n_r$ 

\begin{algorithmic}[1]
\State Initialize critic networks $f_1$, $f_2$ and a actor network $\pi$.
\For{$i = 1,2,\dotsc,N$}
\State Sample minibatch $\cD_{\textrm{mini}}$ from dataset $\cD$.
\State Initialize $\cL(f, \pi), \cE_{\textrm{g}} (f), \cE_{\textrm{td}} (f) = 0$ for each $f\in\{f_1, f_2\}$.
\For{$(s,a,r,s') \in \cD_{\textrm{mini}}$}

\State Generate $\pi_{\textrm{LLM}}(s)$ from LLM responses.
\State Compute losses for each critic $f\in\{f_1, f_2\}$ 
\Statex \qquad \quad $\cL(f, \pi) += f(s, \pi) - f (s, \pi_{\textrm{LLM}})$ 
\Statex \qquad \quad $\cE_{\textrm{g}} (f, \pi_{\textrm{LLM}}) +=  ( f(s,a) - f(s, \pi_{\textrm{LLM}} ))^2$
\Statex \qquad \quad $ \cE_{\textrm{td}} (f, \pi) +=   ( f(s,a) - r - \gamma f(s', a') )^2 $, where $a'\sim \pi(s')$

\EndFor

\State Update critic networks $f\in\{f_1, f_2\}$ \label{ln:update critic}
\Statex\qquad$l_{\textrm{critic}}(f) \coloneqq \frac{1}{|\cD_{\textrm{mini}}|}(
    \cL (f, \pi ) + \alpha \cE_{\textrm{g}} (f, \pi_{\textrm{LLM}}) + \beta \cE_{\textrm{td}} (f, \pi) )$ 
\Statex \qquad$f \gets f - \eta_{\textrm{critic}}\nabla l_{\textrm{critic}} (f, \pi)$ 
\State Update actor network $\pi$ \label{ln:update actor}
\Statex \qquad$l_{\textrm{actor}}(\pi) = - \frac{1}{|\cD_{\textrm{mini}}|} \cL(f_1,\pi)$
\Statex \qquad$\pi \gets \pi - \eta_{\textrm{actor}}\nabla l_{\textrm{actor}}$
\EndFor
\end{algorithmic}
\end{algorithm}

\renewcommand{\arraystretch}{1.5}
\setlength{\tabcolsep}{1.5pt}
  \begin{table*} 
  \centering 
  \resizebox{0.99\textwidth}{!}{
    \begin{tabular}{llcccccccccccccccccc}
      \toprule
      \multirow{2}{*}{Dataset} &\multirow{2}{*}{Models} & \multicolumn{6}{c}{Accuracy} & \multicolumn{3}{c}{Reward}  & \multicolumn{3}{c}{Diversity} & \multicolumn{3}{c}{Novelty}    
      \\\cmidrule(rl){3-8} \cmidrule(rl){9-11} \cmidrule(rl){12-14} \cmidrule(rl){15-17}
      & & HR@5 & HR@10 & HR@20 & NDCG@5 & NDCG@10 & NDCG@20 & R@5 & R@10 & R@20 & CV@10 & CV@20 & Entropy & NCV@10 & NCV@20 & NC@1 
      \\\hline
      &  $\pi_{\mathrm{Llama3}}$ & 0.0036 & 0.0064 & 0.0115 & 0.0022 & 0.0031 & 0.0044 & 55 & 99 & 179 & 0.9803 & 0.9803 & 2.2747 & 0.9692 & 0.9692 & 1,731 \\
      &  $\pi_{\mathrm{Claude3}}$ & 0.0061 & 0.0112 & 0.0161 & 0.0039 & 0.0055 & 0.0067 & 93 & 169 & 242 & 0.5318 & 0.5335 & 2.0651 & 0.2989 & 0.3018 & 209 \\ \arrayrulecolor{gray} \cline{2-17} \arrayrulecolor{black}
      \multirow{3}{*}{MovieLens} 
      & GRU4Rec &  0.0401  &  0.0644  &  0.1026 &  0.0261  &  0.0339  &  0.0435 &  622  &  994  &  1,574  &  0.6773  &  0.7350  & 1.5207 &  0.3764  &  0.4782  & 219 \\
      & SMORL & 0.0380 & 0.0620 & 0.0994 & 0.0243 & 0.0320 & 0.0414 & 587 & 954 & 1,523 & 0.6682 & 0.7310  & - & 0.3644 & 0.4727 &  191   \\     
     & \algoname (Llama3) &  \textbf{0.0458}  & \textbf{ 0.0720}  &  \textbf{0.1104} &  0.0303  &  0.0387  &  0.0484  &  711  &  \textbf{1,109}  &  1,687 &  \textbf{0.6899}  &  \textbf{0.7674} & 8.0594    &  \textbf{0.4235}  &  \textbf{0.5464}   & 268 \\
                               & \algoname (Claude3) &  \textbf{0.0458}  & \textbf{0.0720}  &  \textbf{0.1104} &  \textbf{0.0305}  &  \textbf{0.0389}  &  \textbf{0.0486}  &  \textbf{712}  &  \textbf{1,109}  &  \textbf{1,690} &  0.6877  &  0.7646 & \textbf{8.0595}    &  0.4192  &  0.5410   & \textbf{278} \\ \arrayrulecolor{gray}
      \bottomrule
    \end{tabular}}
  \vspace{5pt}
  \caption{Performance comparison of baseline methods (GRU4Rec and SMORL) and \algoname evaluated on the MovieLens dataset across accuracy, reward, diversity, and novelty metrics. Best scores are highlighted in boldface, excluding the standalone LLM policies $\pi_{\mathrm{Llama3}}$ and $\pi_{\mathrm{Claude3}}$. \algoname (Llama3) and \algoname (Claude3) represent \algoname trained with $\pi_{\mathrm{Llama3}}$ and $\pi_{\mathrm{Claude3}}$ reference policies, respectively.}
    \label{table:movie_acc}
  \end{table*}

\paragraph{Grounding the critic function via regularization.} To guarantee compatibility with the target system without direct modification of $\pi_{\textrm{LLM}}$, we introduce the regularization losses $\cE_{\textrm{g}}$ and $\cE_{\textrm{td}}$ in \eqref{eq:llm_adv_opt}.   
\begin{itemize}[leftmargin=*]
\item \textbf{Temporal difference loss ($\cE_{\textrm{td}}$)}: The TD loss enforces Bellman consistency in the critic ($f$), ensuring it learns realistic values for in-sample actions aligning with the dataset $\mathcal{D}$. For frequently observed actions in $\mathcal{D}$, the loss strongly aligns the critic $f$ with actual reward observations, while for actions rarely observed in $\mathcal{D}$, the loss has a smaller effect, allowing the critic $f$ to maintain optimistic values. Figure~\ref{fig:reward_hypothesis} shows how the critic function is shaped by the TD loss for both actions observed in the dataset and novel actions recommended by the LLM in movie recommendation scenarios.
    \item \textbf{Grounding loss ($\cE_{\textrm{g}}$)}: Limiting critic values only for in-sample actions is insufficient, as it may still overestimate values for LLM-suggested items not seen in $\cD$. To address this, we introduce a grounding loss that constrains $f(s,\pi_{\textrm{LLM}})$, the critic values of $\pi_{\textrm{LLM}}$, to stay close to those of in-sample actions. This reduces excessive optimism for unexplored actions and ensures a grounded evaluation based on the reliable values of popular actions.
      
    \end{itemize}
With proper choices of $\alpha$ and $\beta$, the critic $f$ learns to estimate values for novel items that are optimistic yet still grounded within the constraints of the training datasets. Meanwhile, the policy $\pi$ evolves to generate recommendations that strike a balance between exploring novel items and exploiting popular ones. We will analyze the impact of $\alpha$ and $\beta$ on performance in Section~\ref{sec:experiment}.

\subsection{LLM-guided adversarial actor critic (\algoname)}
To solve the optimization problem \eqref{eq:llm_adv_opt}, we present an LLM-guided Adversarial Actor Critic (\algoname), which constructs an LLM policy $\pi_{\textrm{LLM}}$ from LLMs' responses and trains actor ($\pi$) and critic ($f$) networks to optimize their empirical losses computed based on batch datasets and $\pi_{\textrm{LLM}}$.
The complete description of \algoname is provided in Algorithm~\ref{alg:latac}.

\subsubsection{Constructing LLM policy $\pi_{\textrm{LLM}}$}
\label{sec:approx-llm-policy}
Obtaining the exact probability distribution of the LLM policy $\pi_{\textrm{LLM}}$ over a large action space is challenging, and LLMs may suggest actions outside the predefined space. To address this, we extract recommendations $\cA_{r}$ from the LLM by providing a prompt $p(s, \cA_c, n_r)$, which includes a candidate action set $\cA_c \subseteq \cA$, the size of recommendations $n_r$, and the current state $s$. We then construct the LLM policy $\pi_{\textrm{LLM}}$ as uniformly distributed over $\cA_r$. To control prompt and response length, we randomly sample $\cA_c$ from $\cA$ and specify $n_r = |\cA_r|$.

\subsubsection{Adversarial actor critic training}
We present a practical actor-critic approach that iteratively updates actor ($\pi$) and critic ($f$) networks to optimize the adversarial objectives in \eqref{eq:llm_adv_opt}, where the loss is computed based on a sampled mini-batch dataset $\cD_{\textrm{mini}} \subseteq \cD$ and the LLM policy $\pi_{\textrm{LLM}}$. 
During the critic updates (Line~\ref{ln:update critic} in Algorithm~\ref{alg:latac}), the critic networks $f$ are updated to minimize the adversarial loss $\cL(f, \pi)$ added with the regularization terms $\cE_{\textrm{g}} (f)$ and $\cE_{\textrm{td}} (f)$. This increases the values of novel actions recommended by $\pi_{\textrm{LLM}}$ in comparison to the actions of $\pi$.
To avoid the deadly triad issue, we implement the double Q heuristic~\cite{cheng2022adversarially,fujimoto2018addressing,haarnoja2018soft}, using two critic networks ($f_1$ and $f_2$) for loss computation. Once critic networks are chosen, the actor network $\pi$ is updated to maximize the adversarial loss $\cL(f, \pi)$, in contrast to the critic updates,  thereby improving the policy $\pi$ in relation to the LLM policy $\pi_{\textrm{LLM}}$ (Line~\ref{ln:update actor} in Algorithm~\ref{alg:latac}). After $N$ iterations of the updates, the algorithm outputs the actor $\pi$ and the critics $f_1$ and $f_2$.

\renewcommand{\arraystretch}{1.5}
\setlength{\tabcolsep}{1.5pt}
  \begin{table*} 
  \centering 
  \resizebox{0.99\textwidth}{!}{
    \begin{tabular}{llcccccccccccccccccc}
      \toprule
      \multirow{2}{*}{Dataset} &\multirow{2}{*}{Models} & \multicolumn{6}{c}{Accuracy} & \multicolumn{3}{c}{Reward}  & \multicolumn{3}{c}{Diversity} & \multicolumn{3}{c}{Novelty}    
      \\\cmidrule(rl){3-8} \cmidrule(rl){9-11} \cmidrule(rl){12-14} \cmidrule(rl){15-17}
      & & HR@5 & HR@10 & HR@20 & NDCG@5 & NDCG@10 & NDCG@20 & R@5 & R@10 & R@20 & CV@10 & CV@20 & Entropy & NCV@10 & NCV@20 & NC@1 
      \\\hline
      \multirow{4}{*}{\makecell{MovieLens \\ (skewed)}} 
      &GRU4Rec & 0.0387 & 0.0613 & \textbf{0.0962} & 0.0251 & 0.0324 & 0.0412 & 606 & 953 & \textbf{1,485}  & 0.6460 & 0.6913  & 1.6640 & 0.3433 & 0.4117 &  202 \\
      &SMORL & 0.0330 & 0.0535 & 0.0850 & 0.0215 & 0.0281 & 0.0360 & 515 & 830 & 1,314 & 0.6366 & 0.6900  & - & 0.3312 & 0.4107 &  169   \\     
      &\algoname (Llama3) &  \textbf{0.0399} & \textbf{0.0618} & 0.0932  & 0.0267 & \textbf{0.0337} & \textbf
                                                                                                                                                                                                            {0.0416} & \textbf{624} & \textbf{958} & 1,438 & \textbf{0.6722} & \textbf{0.7444} & 8.0545 & \textbf{0.4041} & \textbf{0.5147} &  264 \\
      &\algoname (Claude3) &  0.0397 & 0.0610 & 0.0929  & \textbf{0.0268} & 0.0336 & \textbf{0.0416} & 621 & 949 & 1,431 & 0.6672 & 0.7397 & \textbf{8.1613} & 0.3959 & 0.5062 &  \textbf{265} \\
      \bottomrule
    \end{tabular}}
  \vspace{5pt}
    \caption{Performance comparison between baseline methods (GRU4Rec and SMORL) and \algoname on the skewed MovieLens dataset, which contains exclusively male user samples. All methods are evaluated on the MovieLens dataset with its original user distribution across metrics of accuracy, reward, diversity, and novelty. Boldface indicates best performance. \algoname (Llama3) and \algoname (Claude3) represent \algoname variants using $\pi_{\mathrm{Llama3}}$ and $\pi_{\mathrm{Claude3}}$ reference policies, respectively.}
    \label{table:movie_skewed}
  \end{table*}

\section{Experiment} 
\label{sec:experiment}

In this section, we evaluate the performance of \algoname and baseline algorithms for recommendation on a real-world datasets.

\subsection{Experimental settings}

\subsubsection{Datasets}
We perform experiments using the real-world movie rating dataset MovieLens-1M \cite{harper2015movielens}, including 1 million ratings for 3,503 movies.
Items with fewer than three interactions and users whose interaction length is smaller than three are removed from the datasets.
From the datasets, we randomly sample ratings, organize them chronologically, and group them by user to create a sequence of interactions for each individual. For each user sequence, at each time step $t$, we define a state as $s_t = G(x_{t-5:t}) $, where $x_{t-5:t}$ represents the five most recent items the user has watched prior to time step $t$, and $G$ is an encoder for the sequential model that transforms the input sequence into a hidden state. Additionally, action $a_t$ corresponds to an item ID rated by the user, and reward $r_t$ is the rating given by the user for that item, which ranges from $[1,5]$. Accordingly, we obtain 26,511 samples from 160 users in the MovieLens dataset, along with the corresponding suggestions generated from LLMs. Of these, 21,421 samples are used for training, while 5,090 samples are reserved for evaluation.

\subsubsection{Baselines}
\begin{itemize}[leftmargin=*]
    \item LLM policy ($\pi_{\textrm{LLM}}$): We first establish the LLM policy as our primary baseline, which represents the untuned performance of LLMs on our task. To ensure our evaluation is robust across different LLMs, we use two distinct LLMs to construct the LLM policy: LLama3-8B-Instruct~\cite{llama3modelcard} and Claude3 Haiku~\cite{claude3modelcard}. The LLM policy operates by generating responses to prompts that describe states in text format (as detailed in Section~\ref{sec:approx-llm-policy}), then creating a uniform distribution over the items suggested in these responses. We denote the resulting baseline policies as $\pi_{\mathrm{Llama3}}$ and $\pi_{\mathrm{Claude3}}$ respectively.
    \item GRU4Rec \cite{hidasi2015session}: 
    GRU4Rec is a recurrent neural network model for RS, utilizing gated recurrent units (GRUs) \cite{cho2014properties}  to encode sequences of user-item interactions. It is trained in a supervised manner using cross-entropy loss to predict the next item.
    \item Scalarized multi-objective RL (SMORL) \cite{stamenkovic2022choosing}: SMORL  is a multi-objective RL approach designed to improve diversity and novelty in RS.  It performs scalarized Q-learning for three different objectives -- accuracy, diversity, and novelty -- by employing distinct output layers for each of these objectives.
\end{itemize}

\usetikzlibrary{patterns}

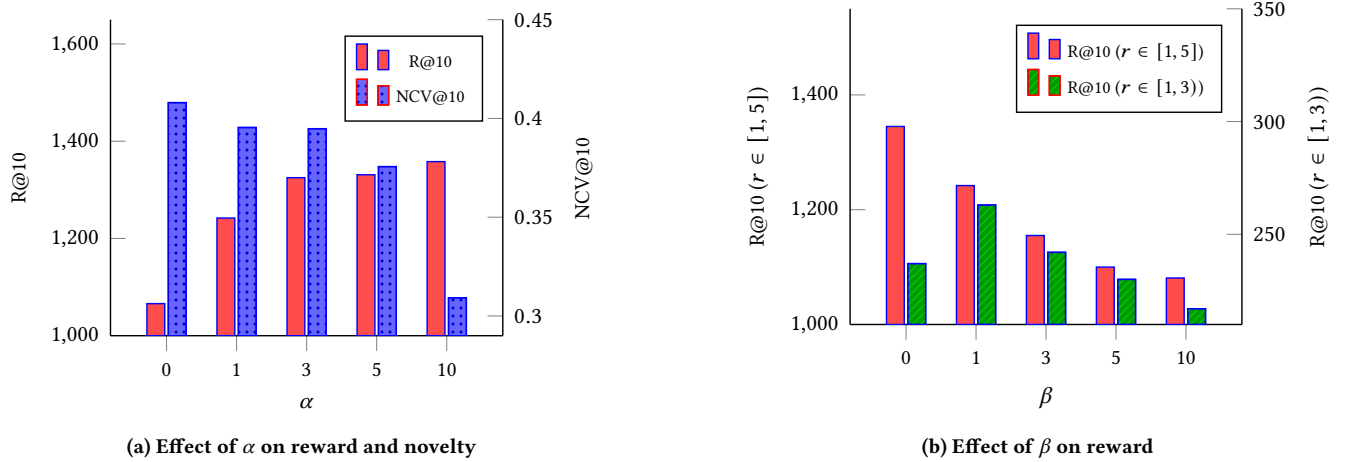
\begin{figure*}[t]
\centering
\begin{subfigure}[t]{0.45\textwidth}
\centering
\begin{adjustbox}{width=\textwidth}
\begin{tikzpicture}
\begin{axis}[
    ybar,
    bar width=5pt,
    enlarge x limits=0.2,
    symbolic x coords={0,1,3,5,10},
    xtick=data,
    height=4.7cm,
    xlabel={$\alpha$},
    ylabel={R@10},
    ymin=1000, ymax=1650,
    axis y line*=left,
    axis x line*=bottom,
    tick label style={font=\scriptsize},
    label style={font=\scriptsize},
    legend style={
        at={(0.95,0.95)},
        anchor=north east,
        legend columns=1,
        font=\tiny
    }
]
\addplot+[ybar, fill=red!70!white, bar shift=-3pt] coordinates {
    (0,1066) (1,1242) (3,1325) (5,1331) (10,1358)
};
\addlegendentry{R@10}
\addplot+[ybar, fill=blue!60!white, bar shift=3pt, postaction={pattern=dots, pattern color=blue!80!black}] coordinates {
    (0,0) (1,0) (3,0) (5,0) (10,0)
}; 
\addlegendentry{NCV@10}
\end{axis}
\begin{axis}[
    ybar,
    bar width=5pt,
    enlarge x limits=0.2,
    symbolic x coords={0,1,3,5,10},
    xtick=\empty,
    height=4.7cm,
    axis x line=none,
    axis y line*=right,
    ymin=0.29, ymax=0.45,
    ylabel={NCV@10},
    bar shift=3pt,
    tick label style={font=\scriptsize},
    label style={font=\scriptsize}
]
\addplot+[ybar, fill=blue!60!white, postaction={pattern=dots, pattern color=blue!80!black}] coordinates {
    (0,0.4080) (1,0.3954) (3,0.3948) (5,0.3756) (10,0.3092)
};
\end{axis}
\end{tikzpicture}
\end{adjustbox}
\caption{Effect of $\alpha$ on reward and novelty}
\label{fig:alpha-effect}
\end{subfigure}%
\hfill%
\begin{subfigure}[t]{0.45\textwidth}
\centering
\begin{adjustbox}{width=\textwidth}
\begin{tikzpicture}
\begin{axis}[
    ybar,
    bar width=5pt,
    enlarge x limits=0.2,
    symbolic x coords={0,1,3,5,10},
    xtick=data,
    height=4.7cm,
    xlabel={$\beta$},
    ylabel={R@10 ($r \in [1,5]$)},
    ymin=1000, ymax=1550,
    axis y line*=left,
    axis x line*=bottom,
    tick label style={font=\scriptsize},
    label style={font=\scriptsize},
    legend style={
        at={(0.95,0.95)},
        anchor=north east,
        legend columns=1,
        font=\tiny
    }
]
\addplot+[ybar, fill=red!70!white, bar shift=-3pt] coordinates {
    (0,1345) (1,1242) (3,1155) (5,1100) (10,1081)
};
\addlegendentry{R@10 ($r \in [1,5]$)}
\addplot+[ybar, fill=green!60!black, bar shift=3pt, postaction={pattern=north east lines, pattern color=green!80!black}] coordinates {
    (0,0) (1,0) (3,0) (5,0) (10,0)
}; 
\addlegendentry{R@10 ($r \in [1,3)$)}
\end{axis}
\begin{axis}[
    ybar,
    bar width=5pt,
    enlarge x limits=0.2,
    symbolic x coords={0,1,3,5,10},
    xtick=\empty,
    height=4.7cm,
    axis x line=none,
    axis y line*=right,
    ymin=210, ymax=350,
    ylabel={R@10 ($r \in [1,3)$)},
    bar shift=3pt,
    tick label style={font=\scriptsize},
    label style={font=\scriptsize}
]
\addplot+[ybar, fill=green!60!black, postaction={pattern=north east lines, pattern color=green!80!black}] coordinates {
    (0,237) (1,263) (3,242) (5,230) (10,217)
};
\end{axis}
\end{tikzpicture}
\end{adjustbox}
\caption{Effect of $\beta$ on reward}
\label{fig:beta-effect}
\end{subfigure}%
\caption{
Performance analysis of \algoname (Llama3) on MovieLens dataset. 
\textbf{Left:} Evaluated rewards (R@10) and novelty (NCV@10) for varying $\alpha=0,1,3,5,10$. Higher $\alpha$ improves reward but reduces novelty. 
\textbf{Right:} Evaluated rewards (R@10) for varying $\beta=0,1,3,5,10$ on the full dataset ($r \in [1,5]$) and the filtered dataset only consisting of samples with poor ratings ($r \in [1,3)$). Too low $\beta$ dcreases cumulative reward (R) when datasets consist of poor actions with low ratings.
}
\label{fig:alphabeta}
\end{figure*}

\subsubsection{Implementation details.}
For all algorithms, except $\pi_{\mathrm{LLM}}$, to generate state $s_t = G(x_{t-5:t})$, we use the five most recent items' embeddings $x_{t-5:t}$ and GRUs \cite{cho2014properties} to encode the input sequence. The item embedding size and hidden size are both set to 64 for all models. The learning rate is set as $0.005$ for GRU4Rec, $0.01$ for SMORL as suggested in \cite{stamenkovic2022choosing}, and $\eta_{\textrm{critic}}=0.01$ and $\eta_{\textrm{actor}}=0.001$ for \algoname to ensure the critic stably evaluates the policy \cite{borkar1997stochastic,maei2009convergent}. For the RL algorithms, we set the discount factor $\gamma$ to $0.5$ for SMORL as suggested in \cite{stamenkovic2022choosing} and $0.99$ for \algoname. For SMORL, we set equal weights for each reward objective for accuracy, novelty and diversity, i.e., $\bm{w}=[1,1,1]$. For \algoname, the default settings for the regularization coefficients are $\alpha=1.0$ and $\beta=1.0$. We use adaptive gradient descent algorithm ADAM~\cite{kingma2014adam} for updates, and train the models for 10,000 steps with a minibatch size of $128$. 

The LLM policy $\pi_{\mathrm{LLM}}$ can be directly extracted from an LLM by prompting state information in a text format.
To generate $\pi_{\mathrm{LLM}}(s_t)$ for a given state $s_t = x_{t-5:t}$, we provide the titles of five items the user has selected before $t$ in a prompt given to the LLM model.  Additionally, to limit the length of prompts and responses, we provide 100 candidate items randomly sampled from the entire item set and specify the number of items to be recommended in a prompt, i.e., $n_c=100$ and $n_r=10$. 
See Section~\ref{sec:approx-llm-policy} for details.  
  
\subsubsection{Metrics}
\begin{itemize}[leftmargin=*]
    \item \textbf{Accuracy/reward: } We consider two commonly used metrics for their top-$k$ predictions, where $k=5,10$ and $20$: hit ratio (HR$@k$) and normalized discounted cumulative gain (NDCG$@k$)~\cite{jarvelin2002cumulated}. HR$@k$ evaluates whether the ground-truth item appears in the top-$k$ positions of the recommendation list, while NDCG$@k$ is a rank-sensitive metric that assigns higher scores to items positioned at the top of the recommendation list.
To assess the long-term effect of algorithms' item recommendations in terms of high ratings, we measure cumulative reward (R$@k$), defined as the sum of user ratings for the top-$k$ recommendations.
\item \textbf{Diversity/novelty: } We measure item coverage (CV$@k$), novel item coverage (NCV$@k$), and the total count of novel items (NC$@k$). Here, novel items indicate items fall within the 50\% tail in terms of popularity. More specifically, CV$@k$ is a ratio of all items (less popular items) covered by all top-$k$ recommendations of the test sequences, and NCV$@k$ is the  item coverage ratio only computed for novel items that fall within the 50\% tail in terms of popularity.
For GRU4Rec, $\pi_{\mathrm{LLM}}$, and \algoname, learning stochastic policies, we measured the entropy of the policies to assess recommendation diversity, where higher entropy indicates more uniform probability distributions across items and thus greater potential for diverse recommendations.
\end{itemize}

All models were trained 50 times independently using different random seeds, and the final performance metrics were obtained by averaging the results across these trials.

\subsection{Analysis}

\subsubsection{LLM Policy: diversity with misaligned relevance} \label{sec:analysis-llmpolicy}
We assess the performance of the LLM policies,  $\pi_{\textrm{Llama3}}$  and  $\pi_{\textrm{Claude3}}$, which are directly derived from the responses of Llama3 and Claude3 without any additional training. As shown in Table~\ref{table:movie_acc}, while $\pi_{\textrm{Llama3}}$ exhibits notable diversity and novelty in its recommendations, their accuracy and cumulative rewards are low. Although $\pi_{\textrm{Claude3}}$ shows relatively higher accuracy and rewards, it still underperforms compared to other baselines trained on the dataset. This outcome is expected, as the LLMs are not perfectly aligned with user preferences in the dataset, resulting in recommendations of lower quality that fail to effectively meet the needs of target users.

\subsubsection{\algoname: diversity with aligned relevance}
We evaluate the performance of our algorithm \algoname against baseline methods (GRU4Rec and SMORL). As presented in Table~\ref{table:movie_acc}, both \algoname (Llama3) and \algoname (Claude3) trained with  $\pi_{\textrm{Llama3}}$  and  $\pi_{\textrm{Claude3}}$, surpass the baselines in both relevance and diversity/novelty metrics, demonstrating the effectiveness of incorporating the knowledge of LLMs. Remarkably, it achieves both high accuracy and diversity without trade-offs, even though the LLM policies are not perfectly aligned with the target dataset (see Section~\ref{sec:analysis-llmpolicy}). This proves that \algoname effectively utilizes unrefined LLMs directly, resulting in significant savings in time and computational resources. 

\subsubsection{Robustness to skewed datasets}
In RS, the distribution of users in training datasets may differ from that of target users in real applications, making robustness to distribution shifts crucial. To evaluate this, we create a skewed dataset with 14,883 ratings from 90 male users and train \algoname and baseline algorithms on it. As shown in Table~\ref{table:movie_skewed}, \algoname and GRU4Rec perform well in almost all metrics accuracy, diversity, and novelty, while SMORL's diversity performance declines significantly with the skewed dataset.
These results demonstrate that \algoname's learned diversity and novelty handle bias from imbalanced data, generalizing well to different distributions.

\subsubsection{The effect of grounding loss ($\alpha$)}
The grounding loss $\cE_{\textrm{g}}$ constrains the critic values of novel actions suggested by the LLM to align with the values of in-sample actions. As shown in Figure~\ref{fig:alphabeta}, increasing $\alpha$ reduces novelty but improves accuracy. This occurs because large $\alpha$ restricts the optimism toward novel suggestions from the LLM, while promoting stable critic learning by ensuring that out-of-sample actions' values align with those of in-sample actions reliably learned with enough observations, which results in more accurate critic estimates.

\subsubsection{The effect of TD loss ($\beta$)}
The TD loss $\cE_{\textrm{td}}$ regularizes the critic $f$ by enforcing Bellman consistency to learn realistic values for in-sample actions. As shown in Figure~\ref{fig:alphabeta}, increasing $\beta$ reduces accuracy, as it makes action estimates less optimistic. However, a very small $\beta$ harms performance by making the policy blindly mimics actions observed in datasets or suggested by the LLM, ignoring actual rewards. We additionally tested our algorithm on a low-quality dataset (actions with ratings below 3) across different $\beta$ values and found that rewards decrease with $\beta = 0$, demonstrating that too small $\beta$ is detrimental, especially with poor-quality datasets.

\section{Conclusion}
We propose a novel RL method that enhances diversity and novelty in recommendations without compromising accuracy, by leveraging LLM-generated suggestions without the need for expensive LLM fine-tuning. Our approach learns a balanced policy capable of recommending both popular and novel items. Experiments on real-world datasets demonstrate that our method achieves significant gains in diversity, novelty, and accuracy, highlighting its effectiveness and practicality for real-world recommendation systems.

\bibliographystyle{plain}
\bibliography{reference}

\end{document}